\documentclass[letterpaper, 10 pt, conference]{ieeeconf}  % Comment this line out if you need a4paper

\IEEEoverridecommandlockouts                              % This command is only needed if
\usepackage{graphicx}
\usepackage{amsmath}
\usepackage{amssymb}
\usepackage{booktabs}
\usepackage{xcolor}
\usepackage{cite}
\usepackage{pifont}
\usepackage{makecell}
\usepackage{subcaption}
\usepackage{placeins}
\newcommand{\yes}{\textcolor{black!80}{\ding{51}}}
\newcommand{\no}{\textcolor{black!35}{\ding{55}}}
\newcommand{\partialmark}{\textcolor{black!60}{$\circ$}}

\title{\LARGE \bf
Locomotion-Grounded Humanoid Soccer: Task-Gated Reinforcement\\
Learning of a Multi-Directional Kicking Library
}

\author{Abu Hanif Muhammad Syarubany, Jaehyun Jang, Hwanhee Kim,\\
Kyuwon Kim, Seungyeon Ryu, and Chang D. Yoo\\
\normalsize School of Electrical Engineering, KAIST, Daejeon, Republic of Korea% <-this % stops a space
}

\begin{document}

\maketitle
\thispagestyle{empty}
\pagestyle{empty}

%%%%%%%%%%%%%%%%%%%%%%%%%%%%%%%%%%%%%%%%%%%%%%%%%%%%%%%%%%%%%%%%%%%%%%%%%%%%%%%%
\begin{abstract}

Recent humanoid soccer systems make motion tracking the substrate and derive
locomotion from it, typically by steering a motion-reference anchor toward the
ball. This yields strong shooting results, but locomotion is trained only on
the narrow, deterministic command distribution ball approach induces, never
evaluated as a capability in its own right. We invert the stack: a general,
command-conditioned locomotion policy is trained first as the substrate, and
$N$ motion-guided kicking skills are added on top as task-gated layers, so
the reachable gait space is set by the locomotion curriculum rather than any
reference clip. Because every skill starts from and returns to this same
commandable state, locomotion also becomes a composition hub ($O(N)$
transitions rather than $O(N^2)$), and post-strike stabilisation is handed
back to the trained controller rather than scripted per clip. We instantiate
this on a 29-DoF Unitree G1 with seven retargeted
kicking skills spanning $259.5^\circ$ of nominal aim direction, including
lateral, rearward and weak-foot strikes a single forward-facing reference
cannot express, and report shooting accuracy alongside command-tracking,
terrain and push-recovery results \emph{with the full skill library attached},
an axis prior humanoid soccer systems do not report. The library is validated
on hardware across forward, lateral, rearward and commanded approaches.

\end{abstract}

%%%%%%%%%%%%%%%%%%%%%%%%%%%%%%%%%%%%%%%%%%%%%%%%%%%%%%%%%%%%%%%%%%%%%%%%%%%%%%%%
\section{INTRODUCTION}

Soccer stresses two humanoid whole-body control capabilities at once, on one
controller and the same hardware. The first is sustained, general locomotion:
repositioning, retreating, sidestepping, turning in place. The second is
precise, contact-rich striking, whose useful signal is a sub-$10$\,ms impulse
with delayed feedback about where the ball went. Reinforcement learning has
advanced both halves
\cite{haarnoja2024soccer,robonaldo,paid,striker,humanx,reactive,goalkeeper}.
The strongest reported shooting result on a full-size humanoid attains
$0.73$\,m mean target error from $3$\,m on hardware, at peak ball speeds of
$13.10$\,m/s \cite{robonaldo}.

The recipe behind these results is consistent across systems. A human kicking
clip is retargeted to the robot \cite{gvhmr,gmr}; a tracking reward supplies
the whole-body coordination pure task rewards struggle to discover
\cite{deepmimic,amp,beyondmimic}; ball and target rewards are layered on top.
Locomotion enters last and indirectly, by steering the tracking anchor toward
the ball while the tracking reward stays active throughout approach
\cite{robonaldo}. Motion tracking is the \emph{substrate}.

This ordering, whose reward machinery we largely adopt, has two consequences
the literature has not measured. First, locomotion is exercised only during
ball approach, so the training command distribution is whatever that heuristic
emits. The anchor-steering law of \cite{robonaldo} issues no lateral, backward
or in-place-rotation command, and no command-tracking, rough-terrain,
push-recovery or omnidirectional result is reported there or in
\cite{paid,striker}. That much is curriculum, not architecture, since Stage-3
locomotion in \cite{robonaldo} is already reward-driven. What is architectural
is whether the tracking objective is switched off or merely discounted while
locomotion trains (Sec.~\ref{sec:gating}). Either way, general locomotion in a
humanoid soccer policy is neither trained broadly nor measured. That is the
gap this paper closes.

Second, training on a single reference kick bounds the repertoire to one
contact type. RoboNaldo \cite{robonaldo} reports a right-footed bias,
degradation at extreme lateral and high targets, and names multi-skill priors
an open problem. A match calls for a \emph{library}: an inside-foot pass, a
lateral flick, a strike behind the current heading.

\begin{figure}[t]
  \setlength{\abovecaptionskip}{2pt}
  \centering
  \includegraphics[width=\columnwidth]{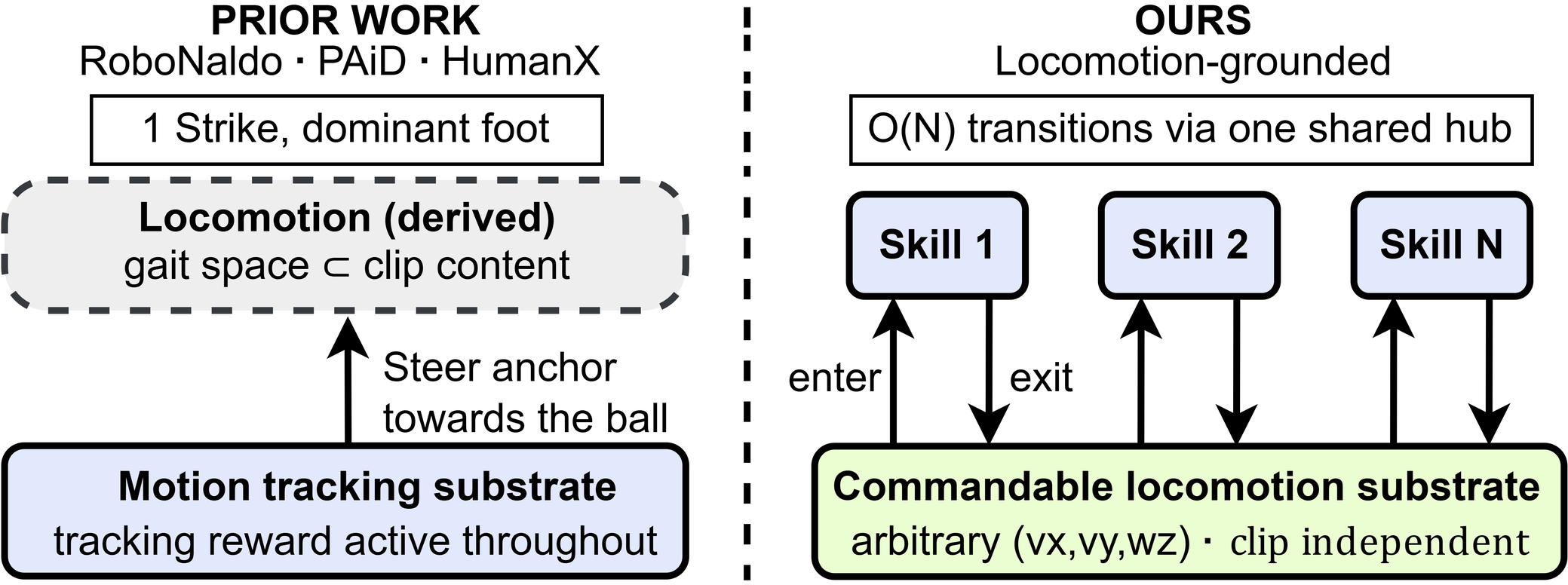}
  \caption{\textbf{Substrate inversion.} Prior work derives locomotion from a
  ball-steered anchor; ours is the commandable base.}
  \label{fig:overview}
\end{figure}

\begin{figure*}[t]
  \centering
  \includegraphics[width=0.98\textwidth]{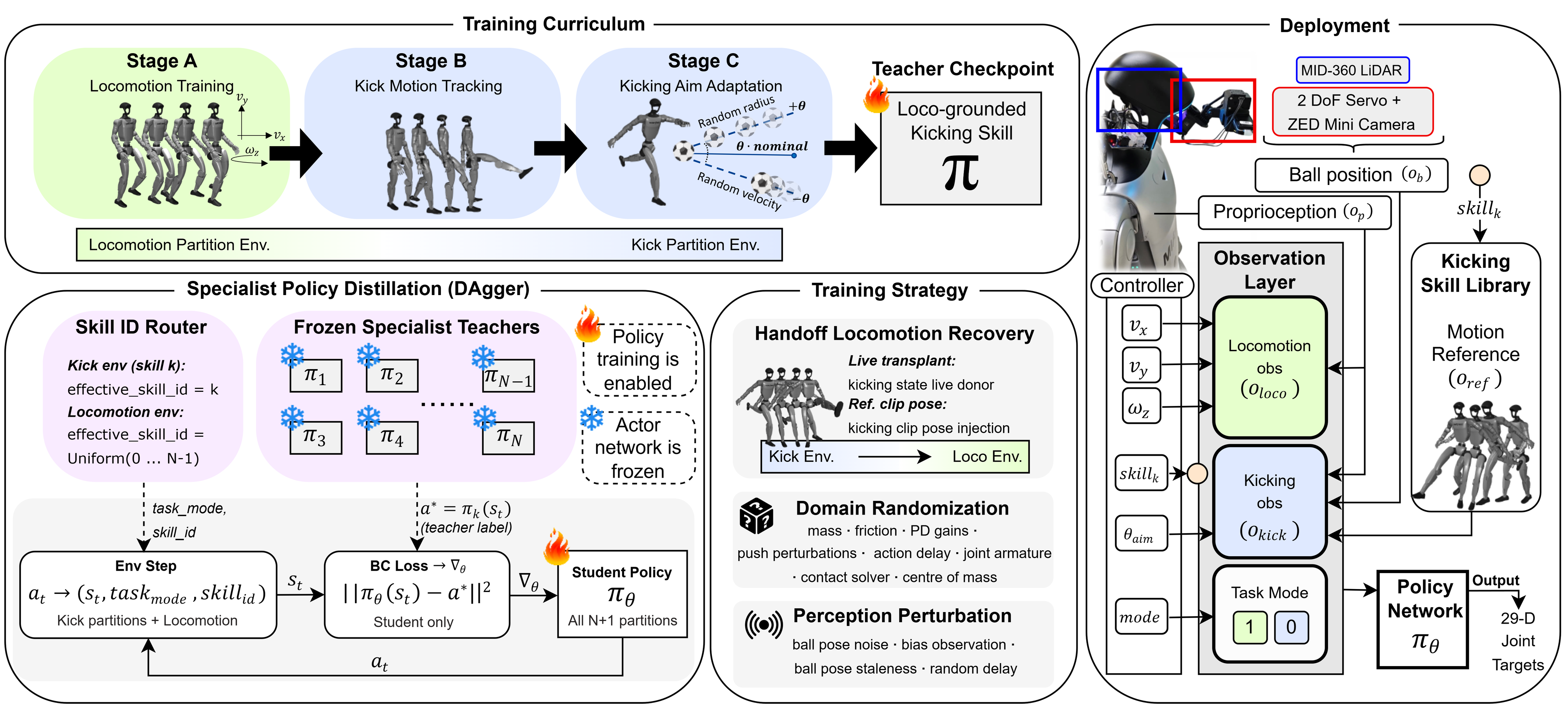}
  \caption{\textbf{Training pipeline and deployed architecture.}
  \textbf{Training (left).} Stages A--C are Stages 1--3 of
  Sec.~\ref{sec:method}: locomotion first and alone, then motion tracking, then
  motion-guided kicking, with specialist distillation (Stage 4) below.
  \textbf{Deployment (right).} The same distilled, task-mode-gated network,
  with the perception hardware of Sec.~\ref{sec:setup} supplying ball
  position.}
  \label{fig:pipeline}
\end{figure*}

\textbf{Approach.} We invert the stack (Fig.~\ref{fig:overview}). A general,
command-conditioned locomotion policy is trained first, with a randomised
$(v_x,v_y,\omega_z)$ envelope, mixed terrain, push/mass/friction/actuation
randomisation, and no ball or reference clip present. $N$ motion-guided
kicking skills are then added as task-gated layers, hard-zeroed in locomotion
mode rather than discounted (Fig.~\ref{fig:pipeline}), so the reachable gait
space is set by the locomotion curriculum alone. Every skill's post-strike
recovery hands back to this same controller, in place of the synthetic,
momentum-blind stabilisation tail prior work relies on: without that tail,
RoboNaldo's alive rate falls $98.8\%{\to}24.4\%$.

We instantiate this on a 29-DoF Unitree G1 at $50$\,Hz. Training runs in
IsaacLab \cite{isaaclab} with an off-policy maximum-entropy learner
\cite{sac,fastsac}, and per-skill specialists are merged into one deployable
network by DAgger-style distillation \cite{dagger,policydistillation}
(Fig.~\ref{fig:pipeline}). We evaluate in MuJoCo \cite{mujoco} sim-to-sim on
the exported policy and on hardware.

\textbf{Contributions.} \textbf{(i) A locomotion-grounded architecture} that
hard-zeroes kick objectives outside kick mode and attaches the skill library
at a locomotion cost of at most $0.04$ in command-tracking MAE
(Table~\ref{tab:loco}). \textbf{(ii) A multi-skill strike library} spanning
$259.5^\circ$ of nominal aim, including lateral, rearward and weak-foot
strikes, reported per-skill and as a union with gaps included. \textbf{(iii) A
learned kick$\to$locomotion handoff} in place of a scripted stabilisation tail,
with an interruptibility analysis of survival when a skill is aborted
mid-swing. \textbf{(iv) A joint evaluation of locomotion generality and
shooting quality}: command tracking, terrain traversal and push recovery with
the library attached, a combination not reported elsewhere.

%%%%%%%%%%%%%%%%%%%%%%%%%%%%%%%%%%%%%%%%%%%%%%%%%%%%%%%%%%%%%%%%%%%%%%%%%%%%%%%%
\section{RELATED WORK}

\subsection{Humanoid Soccer and Robot Sports}

On bipeds, Haarnoja et al.\ \cite{haarnoja2024soccer} demonstrated
end-to-end one-versus-one soccer on a miniature platform, with walking,
turning, kicking and getting up, and transitions learned rather than scripted.
Its policy is conditioned on egocentric game state rather than a velocity
command, and it learns a single scoring strike rather than a library.

On full-size humanoids, a growing line targets shooting directly:
\cite{stoft} optimises foot trajectories without learning; \cite{goalkeeper}
addresses goalkeeping under position-conditioned constraints; \cite{reactive}
and \cite{striker} study vision-driven and noisy-perception striking;
\cite{paid} proposes a progressive perception-action framework; \cite{humanx}
learns generalisable interaction from human video. Concurrently, SkillX \cite{skillx}
composes trapping, dribbling and shooting in one policy, selected by a skill
command rather than a velocity command. It transitions skill-to-skill with no
locomotion state between, and shoots at a goal rather than a commanded aim.
RoboNaldo \cite{robonaldo} is the closest comparison, reporting the strongest
shooting numbers via a three-stage curriculum: clip imitation, ball and target
rewards, then a locomotion command interface. We adopt much of its reward
design. Multi-skill soccer policies exist; what is unreported is locomotion
measured independent of ball approach, and directional strike coverage rather
than a set of distinct behaviours.
Table~\ref{tab:capabilities} scores these axes (\yes\ supported,
\partialmark\ partial, \no\ absent; multi-skill counts
ball-interaction skills, not strike directions). No prior system earns a full
mark on free command, omnidirectional locomotion or rearward targets.

\begin{table*}[htbp]
\centering
\small
\setlength{\tabcolsep}{5pt}
\caption{Capability comparison against prior humanoid soccer systems.
Lateral and rearward targets require the strike bearing to be an explicit
command input spanning that range; systems that reach such bearings only
emergently score \partialmark.}
\label{tab:capabilities}
\begin{tabular}{llcccccccc}
\toprule
& & \multicolumn{2}{c}{\textbf{Locomotion substrate}} & \multicolumn{4}{c}{\textbf{Skill repertoire}} & \multicolumn{2}{c}{\textbf{Handoff}} \\
\cmidrule(lr){3-4}\cmidrule(lr){5-8}\cmidrule(lr){9-10}
Method & Platform & Free & Omni & Multi & Motion & Lateral & Rearward & Loco & Kick \\
 & & command & directional & skill & guided & target & target & $\to$Kick & $\to$Loco \\
\midrule
Haarnoja et al.~\cite{haarnoja2024soccer} & Miniature & \no & \partialmark & \no & \no & \no & \no & \yes & \yes \\
STOFT~\cite{stoft}            & Bipedal  & \no & \no & \yes & \no & \yes & \no & \yes & \yes \\
Reactive~\cite{reactive}      & Humanoid & \no & \no & \partialmark & \yes & \partialmark & \partialmark & \yes & \partialmark \\
Striker~\cite{striker}        & Humanoid & \partialmark & \no & \no & \no & \no & \no & \yes & \yes \\
Goalkeeper~\cite{goalkeeper}  & Humanoid & \no & \no & \yes & \yes & \no & \no & \no & \no \\
HumanX~\cite{humanx}          & Humanoid & \no & \no & \yes & \yes & \no & \no & \no & \no \\
PAiD~\cite{paid}              & Humanoid & \no & \no & \yes & \yes & \no & \no & \yes & \no \\
RoboNaldo~\cite{robonaldo}    & Humanoid & \partialmark & \no & \no & \yes & \partialmark & \no & \yes & \partialmark \\
SkillX~\cite{skillx}          & Humanoid & \no & \no & \yes & \yes & \partialmark & \no & \no & \no \\
\midrule
\textbf{Ours}                 & Humanoid & \yes & \yes & \yes & \yes & \yes & \yes & \yes$^{\S}$ & \yes \\
\bottomrule
\end{tabular}

{\footnotesize $^{\S}$Survival stays above $95\%$ while entry accuracy falls
with the speed carried into the strike (Sec.~\ref{sec:handoff-results}).}

\end{table*}

\subsection{Motion Tracking, Locomotion and Object Interaction as Primitives}

Reference-guided RL supplies whole-body coordination that task rewards alone
struggle to discover \cite{deepmimic,amp}. Recent humanoid systems track
retargeted motion directly on hardware \cite{beyondmimic,gvhmr,gmr}, the
pipeline we use for every clip. Tracking supplies stability but constrains
timing and contact selection, a tension staged curricula aim to relax
\cite{robonaldo}. Our task-mode gating relaxes it fully, switching the
tracking objective off rather than discounting it outside the skill.
Separately, massively parallel simulation has made velocity-commanded
locomotion with terrain and disturbance curricula well established
\cite{isaaclab}, recently at minutes of wall-clock cost via
off-policy maximum-entropy learning \cite{sac,fastsac}. We place this at the
\emph{base} of the stack, not as a ball-approach by-product. The contribution is learner-agnostic: task
gating, the hub and the handoff are properties of the formulation, not the
optimiser. Contact-rich loco-manipulation usually assumes dwell-time contact, with
reward accumulated over many steps \cite{falcon}. Shooting is the opposite: a
brief impulse whose reward arrives only after the ball has travelled. That
motivates the aiming rewards used here and in \cite{robonaldo}, and explicit
handling of post-contact stability.
\subsection{Skill Libraries, Composition and Transitions}

The options framework formalises temporally extended actions via initiation
sets, policies and termination conditions \cite{options}. A shared hub state
makes every skill's termination lie in the next's initiation by construction,
reducing transitions from $O(N^2)$ to $O(N)$. That is the role our locomotion
mode plays. We merge specialists into one network via behaviour distillation
\cite{dagger,policydistillation}: independently trained teachers supervise one
student under its own state distribution. Haarnoja et al.\ merge differently, training under self-play with a KL
regulariser that keeps the policy near the skills; ours regresses directly
onto a routed teacher's actions, with no reward term in merging. A single
deployed network is not itself distinguishing: both
\cite{haarnoja2024soccer} and \cite{robonaldo} deploy one non-hierarchical
network. What differs is which capability it is grounded in, and how many
strike types it carries.

%%%%%%%%%%%%%%%%%%%%%%%%%%%%%%%%%%%%%%%%%%%%%%%%%%%%%%%%%%%%%%%%%%%%%%%%%%%%%%%%
\section{PROBLEM FORMULATION}

We formulate the problem as a single task-conditioned Markov decision process
$(\mathcal{S}, \mathcal{A}, P, r, \gamma)$ solved by one policy $\pi_\theta$.
Each environment additionally carries a task-mode variable
\begin{equation}
m_t \in \mathcal{M} = \{\textsc{loco}, \textsc{kick}_1, \dots, \textsc{kick}_N\},
\label{eq:taskmode}
\end{equation}
selecting, at every timestep, which reward and termination specification the
transition is evaluated against. This is the mechanism behind the \emph{free
command} and \emph{omnidirectional} columns, where
Table~\ref{tab:capabilities} gives no prior system a full mark. $\mathcal{M}$ contains \textsc{loco} as an
ordinary mode, not a residual of ball approach.

\subsection{Task-mode gating}
\label{sec:gating}

Rather than arbitrate between separate controllers, $m_t$ gates reward,
termination and observation inside one network. With $\mathcal{T}$ indexing
reward terms and $w_\tau$ the weight of term $\tau$,
\begin{equation}
r_t = \sum_{\tau \in \mathcal{T}} g_\tau(m_t)\, w_\tau\, \hat{r}_\tau(s_t, a_t),
\qquad
g_\tau : \mathcal{M} \to \{0, 1\},
\label{eq:gating}
\end{equation}
an indicator equal to $1$ only in the mode $\tau$ was authored for. The gate
is a hard zero, not a reduced weight: in locomotion mode every kick term
contributes nothing to the gradient, and conversely. This is what carries the
substrate claim, and Table~\ref{tab:ablations}(c) measures it. Discounting the
tracking terms rather than zeroing them costs command tracking, and the cost
grows with the discount: backward MAE rises $0.12$ at a $0.1\times$ floor,
the weight \cite{robonaldo} itself applies in locomotion mode, and $0.33$ at
$0.5\times$, the latter worse at all eight matched checkpoints (sign test
$p{=}0.008$). All four command axes degrade monotonically. Terminations are
gated identically.

The observation is gated by masking, not omission: inactive-mode terms are
zeroed but stay in the vector, so the input is $261$-dimensional in every mode
(Table~\ref{tab:obs}). A loco/kick one-hot is never masked, so the policy
is told its mode rather than inferring it from which blocks are zero.

\subsection{Skill conditioning}

Within kick mode a skill is specified two ways. Its retargeted reference
trajectory enters the observation directly (Table~\ref{tab:obs}), making the
policy reference-conditioned rather than index-conditioned: there is no
learned skill embedding. The strike direction is \emph{commanded}, not
targeted directly as a world-frame point. Given the ball's realised planar
position $p_{\mathrm{ball}}$ (itself randomised at reset) and a skill's
nominal bearing $\beta$, the target is synthesised as
\begin{equation}
p_{\mathrm{tgt}} = p_{\mathrm{ball}} + D\,[\cos(\beta+\theta),\ \sin(\beta+\theta)]^\top,
\label{eq:target}
\end{equation}
with $D{=}5$\,m for every skill and $\theta$ sampled once per attempt. Only
$\theta$ reaches the policy, normalised as
\begin{equation}
\theta_{aim} = \theta / \theta_{\mathrm{ref}} \in [-1, 1],
\label{eq:aim}
\end{equation}
with $\theta_{\mathrm{ref}}{=}45^\circ$ a fixed constant, so the command is bounded by
construction and needs no robot localisation at deployment. It occupies a
two-wide observation slot whose second component is reserved and held at
zero (Table~\ref{tab:obs}).

Eq.~\ref{eq:target} is what separates this from targeting a fixed
world-frame point, as in \cite{robonaldo}'s $8\times2$\,m goal plane $5$\,m
ahead. Wherever reset randomisation placed $p_{\mathrm{ball}}$,
$p_{\mathrm{tgt}} - p_{\mathrm{ball}}$ is exactly $D$ in direction
$\beta+\theta$, so $\theta$ alone carries the aim error. A world-frame point
target cannot give this: its scale and reachable range are set by each skill's
own geometry. Here $\theta_{aim}\in[-1,1]$ is one bounded command axis for
all $N$.

\subsection{Handoff}

A skill ends when its authored reference content is exhausted, $m_t$ returns
to \textsc{loco}, and the trained locomotion policy resumes control. We call this a handoff rather than a stabilisation phase because no synthetic
recovery reference is tracked. The robot is handed back to a
disturbance-robust controller in whatever single-support, momentum-carrying
state the strike left it. The reverse transition,
Loco$\to$Kick, is commanded rather than learned, with the precondition
Table~\ref{tab:capabilities} footnotes: entry accuracy falls with the velocity
carried into the strike. The simulated sweep also places the ball rather than
walking to it (Sec.~\ref{sec:handoff-results}).

Table~\ref{tab:obs} details the observation space of Fig.~\ref{fig:pipeline}'s
deployment view; every term is robot-measurable except base linear velocity,
which is critic-only.

\begin{table}[htbp]
\centering
\caption{Actor observation space, $261$ dimensions.}
\label{tab:obs}
\footnotesize
\setlength{\tabcolsep}{4pt}
\begin{tabular}{@{}lc@{\hskip 12pt}lc@{}}
\toprule
\multicolumn{2}{c}{\emph{Locomotion group} ($o_{loco}$)} &
\multicolumn{2}{c}{\emph{Kicking group} ($o_{kick}$)} \\
\cmidrule(r){1-2}\cmidrule(l){3-4}
Term & Dim & Term & Dim \\
\midrule
Base angular velocity & $3$  & Motion reference $(q,\dot{q})$ & $58$ \\
Projected gravity     & $3$  & Reference orientation          & $6$ \\
Command $(v_x, v_y, \omega_z)$ & $3$ & Base angular velocity   & $3$ \\
Joint position        & $29$ & Joint position                 & $29$ \\
Joint velocity        & $29$ & Joint velocity                 & $29$ \\
Previous action       & $29$ & Previous action                & $29$ \\
Gait phase, $\sin$    & $2$  & Ball position $o_b$            & $3$ \\
Gait phase, $\cos$    & $2$  & Aim command $\theta_{aim}$     & $2$ \\
\midrule
\multicolumn{4}{@{}l}{Task-mode one-hot, active in both modes: $2$} \\
\bottomrule
\end{tabular}
\end{table}

\section{METHOD}
\label{sec:method}

Training proceeds in the four stages of Fig.~\ref{fig:pipeline}. The ordering is
the claim: locomotion is trained first and alone, and everything else is layered
onto it.

\subsection{Stage 1: locomotion substrate}

A command-conditioned locomotion policy is trained with no ball and no reference
clip present, on commands sampled from a randomised $(v_x, v_y, \omega_z)$
envelope covering forward, backward, lateral and in-place rotation, over mixed
terrain with disturbance randomisation. This stage fixes the reachable gait
space, and with no clip present that space is a property of the command
curriculum alone.

\subsection{Stage 2: motion tracking}

Kick-partition environments add a tracking objective against a human kick clip,
recovered from video \cite{gvhmr} and retargeted \cite{gmr}. Ball and target
rewards stay at zero weight here, so the policy learns the strike's
coordination before it is asked to aim. Locomotion environments continue the
unmodified Stage-1 task: this is an addition, not a replacement.

\subsection{Stage 3: motion-guided ball kicking}
\label{sec:stage3}

The shooting objective ramps linearly from zero over $5000$ iterations, held
at $0$ throughout Stage 2. The aim angle $\theta$ of Eq.~\ref{eq:target} is
sampled per attempt within $\pm15^\circ$ of the skill's nominal bearing, with
ball placement randomised by $\pm10$\,cm; Table~\ref{tab:rewards} lists the main
activated reward terms (each group additionally scaled: tracking $2\times$,
shooting $0.5\times$, others $1\times$), with the alive reward attenuated
$10\times$ before contact so standing still is not viable.

\begin{table}[t]
\centering
\caption{Kick-mode reward terms and weights. Twenty-two standard safety and
regularisation terms (weights $-100$ to $50$) are omitted for space.}
\label{tab:rewards}
\scriptsize
\setlength{\tabcolsep}{3pt}
\begin{tabular}{@{}lr@{\hskip 8pt}lr@{}}
\toprule
Term & $w$ & Term & $w$ \\
\midrule
\multicolumn{4}{@{}l}{\emph{Motion tracking}} \\
Global root pos.        & $1.0$  & Global root orient.    & $0.5$ \\
Relative body pos.      & $1.0$  & Relative body orient.  & $1.0$ \\
Global body lin. vel.   & $1.0$  & Global body ang. vel.  & $1.0$ \\
Global feet lin. vel.   & $1.0$  & Strike joint null-sp.  & $1.0$ \\
Strike leg null-sp.     & $1.0$  &                        &       \\
\addlinespace[2pt]
\multicolumn{4}{@{}l}{\emph{Shooting}} \\
Ball-to-target error    & $20.0$ & Ball velocity          & $10.0$ \\
Predicted ball-to-tgt.  & $5.0$  & Contact orientation    & $4.0$ \\
Ball proximity          & $2.0$  & CoM to ball dist.      & $2.0$ \\
Torso to ball dist.     & $2.0$  & Foot strike pitch      & $1.0$ \\
Ball over line          & $1.0$  & Weak foot contact      & $-1.0$ \\
Self contact (feet)     & $-0.4$ &                        &       \\
\addlinespace[2pt]
\multicolumn{4}{@{}l}{\emph{Post-strike recovery posture}} \\
Stand height            & $-40$  & Stance asymmetry       & $-20$ \\
Stand orientation       & $-20$  & Feet width             & $-20$ \\
Knee width              & $-20$  & Yaw drift              & $-2$ \\
\addlinespace[2pt]
\multicolumn{4}{@{}l}{\emph{Alive and balance shaping}} \\
Kick alive              & $4.0$  & Balance shaping (Eq.~\ref{eq:shaping}) & $1.0$ \\
\bottomrule
\end{tabular}

\end{table}

\textbf{Balance-potential shaping.} The nine $\exp(-e^2/\sigma^2)$ tracking terms of
Table~\ref{tab:rewards} saturate toward $0$ exactly where error is diverging toward a fall, and the
flat per-step alive bonus has zero action-gradient by construction, leaving
the pre-fall regime without a usable signal. We add potential-based shaping
over a capture-point balance margin,
\begin{equation}
F(s,s') = \gamma\,\Phi(s') - \Phi(s), \qquad \Phi(s) \in [0,1],
\label{eq:shaping}
\end{equation}
with $\Phi = \Phi_{\mathrm{bal}}\Phi_{\mathrm{ht}}$ a balance margin
(capture-point distance to the support polygon) times a fall-height margin.
Ng et al.~\cite{ng1999shaping} show this form leaves the optimal policy
unchanged for any $\Phi$ and weight, provided $\gamma$ matches the agent's
discount and $\Phi$ is zeroed only at a true termination; unlike every other
weight in Table~\ref{tab:rewards}, which trades stability against kick
quality, this one can be set aggressively, and no analogous shaping is
reported in \cite{robonaldo}'s otherwise-similar rewards.

\begin{table}[htbp]
\centering
\caption{Skill library geometry.}
\label{tab:skills}
\footnotesize
\setlength{\tabcolsep}{3.5pt}
\begin{tabular}{@{}llrc@{}}
\toprule
Skill & Foot & Nominal & Reachable band \\
      &      & bearing & ($\theta_{\max}=\pm15^\circ$) \\
\midrule
skill\_1 & right & $-1.32^\circ$   & $[-16.3, +13.7]$ \\
skill\_2 & right & $+137.80^\circ$ & $[+122.8, +152.8]$ \\
skill\_3 & left  & $-121.74^\circ$ & $[-136.7, -106.7]$ \\
skill\_4 & right & $-56.94^\circ$  & $[-71.9, -41.9]$ \\
skill\_5 & right & $+48.39^\circ$  & $[+33.4, +63.4]$ \\
skill\_6 & right & $0.00^\circ$    & $[-15.0, +15.0]$ \\
skill\_7 & right & $+43.89^\circ$  & $[+28.9, +58.9]$ \\
\midrule
\multicolumn{4}{@{}l}{Nominal span $259.5^\circ$; reachable union $155.8^\circ$ over five islands} \\
\bottomrule
\end{tabular}

\end{table}

\textbf{Handoff boundary training.} Some locomotion environments are seeded
at reset from a kick-disturbed pose, taken from the reference clip or, with
higher probability, a live running kick environment's own state. The latter is
faithful where the former is not, since tracking error is never zero and
contact forces have no motion-capture representation.
The seeding is enabled during distillation (Stage 4); the ablation retrains it at
specialist level to isolate it (Table~\ref{tab:ablations}(b)).

\begin{table*}[t]
\centering
\caption{Shooting quality in context. PPO, AMP and PAiD rows are the
free-kick simulation baselines as reported in \cite{robonaldo}.}
\label{tab:context}
\scriptsize
% 3pt, not the 3.5pt this table used before the Hit column: adding Hit
% (2026-09-12) pushed the tabular 3.68pt past \columnwidth at 3.5pt.
\setlength{\tabcolsep}{2.8pt}
\begin{tabular}{@{}llcrrrrrr@{}}
\toprule
Method & Bearing range & Dist.\,(m), tol. & suc.@$0.5$\,m (\%) & suc.@$1.0$\,m (\%) & Shot err.\ (m) & $v^{\max}_{\mathrm{ball}}$ (m/s) & Hit (\%) & Alive (\%) \\
\midrule
PPO \cite{ppo}                                   & forward & n/r & $0.0{\pm}0.0$  & $0.0{\pm}0.0$  & $4.721{\pm}.007$ & $0.780{\pm}.003$  & $0.0{\pm}0.0$  & $0.0{\pm}0.0$ \\
AMP \cite{amp}                                   & forward & n/r & $0.9{\pm}0.1$  & $3.3{\pm}0.1$  & $3.733{\pm}.004$ & $1.771{\pm}.051$  & $38.6{\pm}2.8$ & $41.6{\pm}2.3$ \\
PAiD \cite{paid}                                 & forward & n/r & $8.2{\pm}0.7$  & $19.5{\pm}0.5$ & $1.850{\pm}.026$ & $4.986{\pm}.017$  & $67.7{\pm}0.3$ & $67.3{\pm}0.5$ \\
RoboNaldo \cite{robonaldo}, stage 2, free-kick    & forward, $\pm38.7^\circ$ & $5.0$, n/r & $28.8{\pm}1.3$ & $65.5{\pm}0.6$ & $0.899{\pm}.012$ & $\mathbf{14.792{\pm}.011}$ & $82.1{\pm}0.3$ & $\mathbf{100.0{\pm}0.0}$ \\
RoboNaldo \cite{robonaldo}, stage 3, free-kick    & forward, $\pm38.7^\circ$ & $5.0$, n/r & $36.3{\pm}0.1$ & $66.5{\pm}0.2$ & $1.055{\pm}.012$ & $14.178{\pm}.034$ & $81.1{\pm}0.1$ & $99.7{\pm}0.1$ \\
RoboNaldo \cite{robonaldo}, stage 3, moving-ball  & forward, $\pm38.7^\circ$ & $5.0$, n/r & $32.4{\pm}0.6$ & $63.3{\pm}0.6$ & $1.131{\pm}.004$ & $13.875{\pm}.004$ & $79.2{\pm}0.1$ & $98.8{\pm}0.1$ \\
\textbf{Ours}, loco-grounded specialist, skill\_1 & $\approx$forward, $\pm15^\circ$ & $5.0$, $\pm5.7^\circ$ & $\mathbf{44.3{\pm}16.8}$ & $\mathbf{77.5{\pm}12.1}$ & $\mathbf{0.69{\pm}.17}$ & $3.67{\pm}.15$ & $\mathbf{100.0{\pm}0.0}$ & $\mathbf{100.0{\pm}0.0}$ \\
\midrule
\multicolumn{9}{@{}l}{\itshape Multi-directional library: a broader capability envelope (Table~\ref{tab:capabilities})}\\
\textbf{Ours}, unified $N{=}7$, library  & \makecell[l]{$259.5^\circ$ span,\\$155.8^\circ$ union} & $5.0$, $\pm5.7^\circ$ & $15.4{\pm}12.7$ & $32.2{\pm}21.9$ & $1.68{\pm}.76$ & $2.86{\pm}.77$ & $96.7{\pm}5.9$ & $99.4{\pm}1.0$ \\
\bottomrule
\end{tabular}

\end{table*}

\subsection{Stage 4: specialist distillation}

Each skill is trained as above from the shared Stage-1 substrate, one
locomotion-grounded specialist per clip. The specialists are then merged into
one deployable actor $\pi_\theta$ by DAgger-style distillation
\cite{dagger,policydistillation}:
\begin{equation}
\min_\theta\ \mathbb{E}_{s \sim d_{\pi_\theta}}\big[\,\mathcal{L}\big(\pi_\theta(s),\, \pi^{(k(s))}(s)\big)\,\big],
\label{eq:dagger}
\end{equation}
where $k(s)$ is the skill-index router selecting which frozen teacher labels
state $s$: a kick-partitioned environment keeps its own specialist through its
locomotion ticks, and a purely locomotion-partitioned one draws a skill index
uniformly over the $N$ specialists. $d_{\pi_\theta}$ is the state distribution
induced by the \emph{student's own} rollout, not the teachers' trajectories,
which is what separates Eq.~\ref{eq:dagger} from behaviour cloning:
supervision covers the states the student drifts into.

Two details are load-bearing: each teacher must run under its own frozen
observation normaliser (the wrong one yields a confident wrong action and
raises nothing), and the rollout action form must match the regression
target, since sampling one and regressing the other is self-amplifying.

\subsection{Learning algorithm}

All stages use FastSAC, an off-policy maximum-entropy actor-critic
\cite{sac,fastsac}, adopted because simulation throughput, not environment
count, is the bottleneck in contact-rich training, so replay extracts more
updates per simulated step. Running two tasks in one network motivates a
mode-dependent discount and entropy target, $\gamma(m_t)$ and $\alpha(m_t)$,
keyed to $m_t$ as in Eq.~\ref{eq:gating}, so the horizon suited to a
multi-second kick is not forced onto locomotion's shorter one.

\subsection{Domain randomisation}

Randomisation broadens once at the Stage-1 to Stage-2 boundary, then holds
constant. Beyond mass, centre-of-mass, friction, PD-gain and action-delay
randomisation, two axes target the strike specifically: joint armature with
internal friction, and the contact solver's own response parameters. Standard
humanoid randomisation leaves how contact is \emph{resolved} fixed throughout,
which a policy can overfit for free. The ball observation is perturbed on its
own axes (additive noise, a fixed per-episode bias, delayed and held
samples, a stale-reading probability), so it never sees a perfect estimate.

\section{EXPERIMENTAL SETUP}
\label{sec:setup}

\subsection{Platform, simulation and skill library}

All experiments use a 29-DoF Unitree G1, policy at $50$\,Hz emitting joint
position targets, trained in IsaacLab \cite{isaaclab} with thousands of parallel
environments. Every simulated performance number comes from the exported
deployment policy running in MuJoCo \cite{mujoco}, not from training telemetry,
so the cross-engine transfer gap sits inside the measurement rather than behind
it. On hardware, ball position is supplied onboard by a head-mounted MID-360
LiDAR and a $2$-DoF servo-mounted ZED Mini camera
(Fig.~\ref{fig:pipeline}).

Seven kicking skills, each retargeted from a separate human video clip, are
summarised in Table~\ref{tab:skills}: each nominal bearing follows from the
skill's configured ball and target placement ($0^\circ$ straight ahead,
positive left), and the reachable band is that bearing
$\pm\,\theta_{\max}$, with $\theta_{\max}=15^\circ$.

\begin{figure*}[t]
  \centering
  \includegraphics[width=0.82\textwidth]{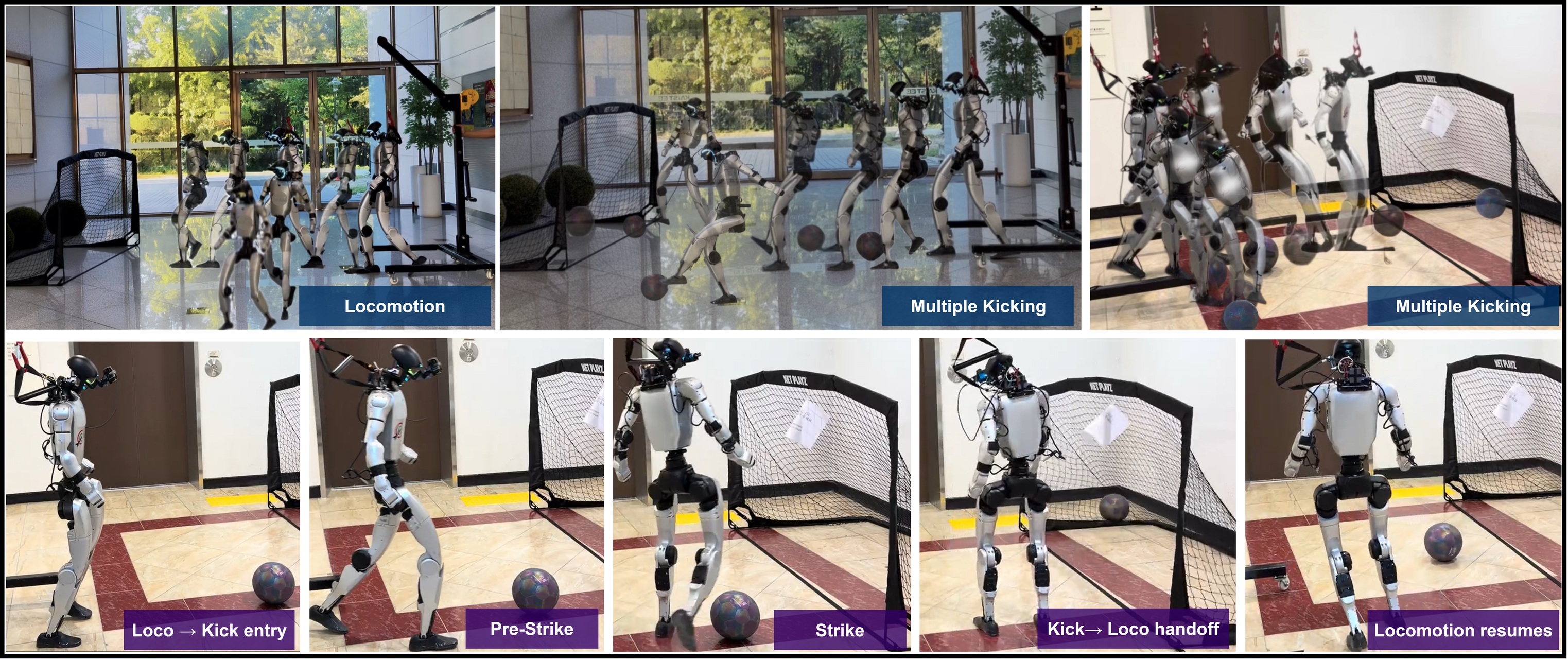}
  \caption{\textbf{Hardware experiments on the Unitree G1.}
  \textbf{Top:} commanded locomotion and two library strikes, each a
  multi-exposure composite of one trial. \textbf{Bottom:} a kick-to-locomotion
  handoff, from the mode switch through contact to locomotion
  resuming with no scripted stabilisation tail (Sec.~\ref{sec:handoff-results}).}
  \label{fig:hardware-setup}
\end{figure*}

\subsection{Metrics and protocol}

Shooting success is scored at $0.5$\,m and $1.0$\,m radius (nearest
post-contact ball-to-target distance; a whiff counts as failure), alongside
shot error, peak ball speed and contact rate. With every target at $D{=}5$\,m
(Eq.~\ref{eq:target}), a $0.5$\,m success allows $\pm5.7^\circ$ of aim for
every skill. Stability uses a
strict fall criterion, not the tracking-based alive flag. Locomotion
generality is command-tracking error, terrain traversal and push recovery,
all with the library attached.

Unified-policy numbers are medians over a checkpoint sweep ($8$ checkpoints,
$16$ for $N{=}7$, $250$--$300$ attempts each), not single readings:
single-checkpoint evaluation is unreliable (one skill ranged
$3.6$--$45.2\%$ across a $16$-checkpoint sweep). Teacher \emph{success} is not
swept, since each teacher is the fixed distillation artifact. We report
medians, not means, since per-checkpoint distributions are zero-heavy and
right-skewed. All results are single training runs except
Table~\ref{tab:loco}'s baseline, which is measured over three seeds.

\section{RESULTS}

Teacher entries are the single checkpoint each student was distilled from;
every other number follows Sec.~\ref{sec:setup}.

\subsection{Shooting quality}
\label{sec:hardware}

Table~\ref{tab:context} compares against prior systems on the metrics they
report (bold = best; n/r = not reported; Hit = ball-foot contact, RoboNaldo's
Contact). The $\pm$ differs by row: seed spread as published for prior
systems, scan spread for our specialist, across-skill spread for our library.
Point accuracy is rarely reported: of the systems in
Table~\ref{tab:capabilities}, only \cite{robonaldo} scores shots by distance
to a target point, the metric we adopt. Within our own table, only the upper, forward-cone
block is like-for-like; the lower, multi-directional row is a different
capability envelope, not a worse entry in the same contest. At comparable
forward distance the skill\_1 specialist reaches $44.3\%$ success at
$0.5$\,m, with standard error $1.7$ over $100$ scans. That is the highest figure we
are aware of on this metric, but it is not a like-for-like win: the aim cone
is $\pm15^\circ$ against $\pm38.7^\circ$, the ball leaves at roughly a
quarter of the pace, and it is one training run. The distilled seven-skill
policy does not retain it: skill\_1 falls to $10.2\%$, and the library mean is
$15.4\%$.

Peak ball speed is the one column prior work dominates, and the gap is an
operating point rather than a ceiling. In the five teachers with training logs,
shooting earns $10$--$15\%$ of the positive kick-mode reward and motion
tracking $76$--$85\%$. Raising the shooting weight several-fold in an
earlier configuration collapsed sim-to-sim survival outright, since strike
power and destabilising contact force are physically coupled, not merely
reward-coupled. The library is tuned for directional
coverage at a survivable strike, not for pace.

Table~\ref{tab:hardware}(a) tests the library on hardware
(Fig.~\ref{fig:hardware-setup}). Skills are grouped by bearing from forward:
skill\_1 and skill\_6 forward, skill\_4 and skill\_5 lateral, skill\_2 and
skill\_3 rearward; skill\_7 overlaps skill\_5 and was not run separately. A
commanded-approach condition exercises the locomotion-to-kick handoff under
operator velocity commands. The test room allows only $1.42$--$2.08$\,m to
the target, against $D{=}5$\,m in simulation. Success therefore scales the
$0.5$\,m radius to the nearest target, $r=0.5\,d_{\min}/D=0.14$\,m with
$d_{\min}=1.42$\,m, so no trial is allowed more than the simulated
$\pm5.7^\circ$.

\subsection{Kicking skill coverage}

\begin{figure}[t]
  \centering
  \begin{subfigure}{0.49\columnwidth}
    \centering
    \includegraphics[width=\linewidth]{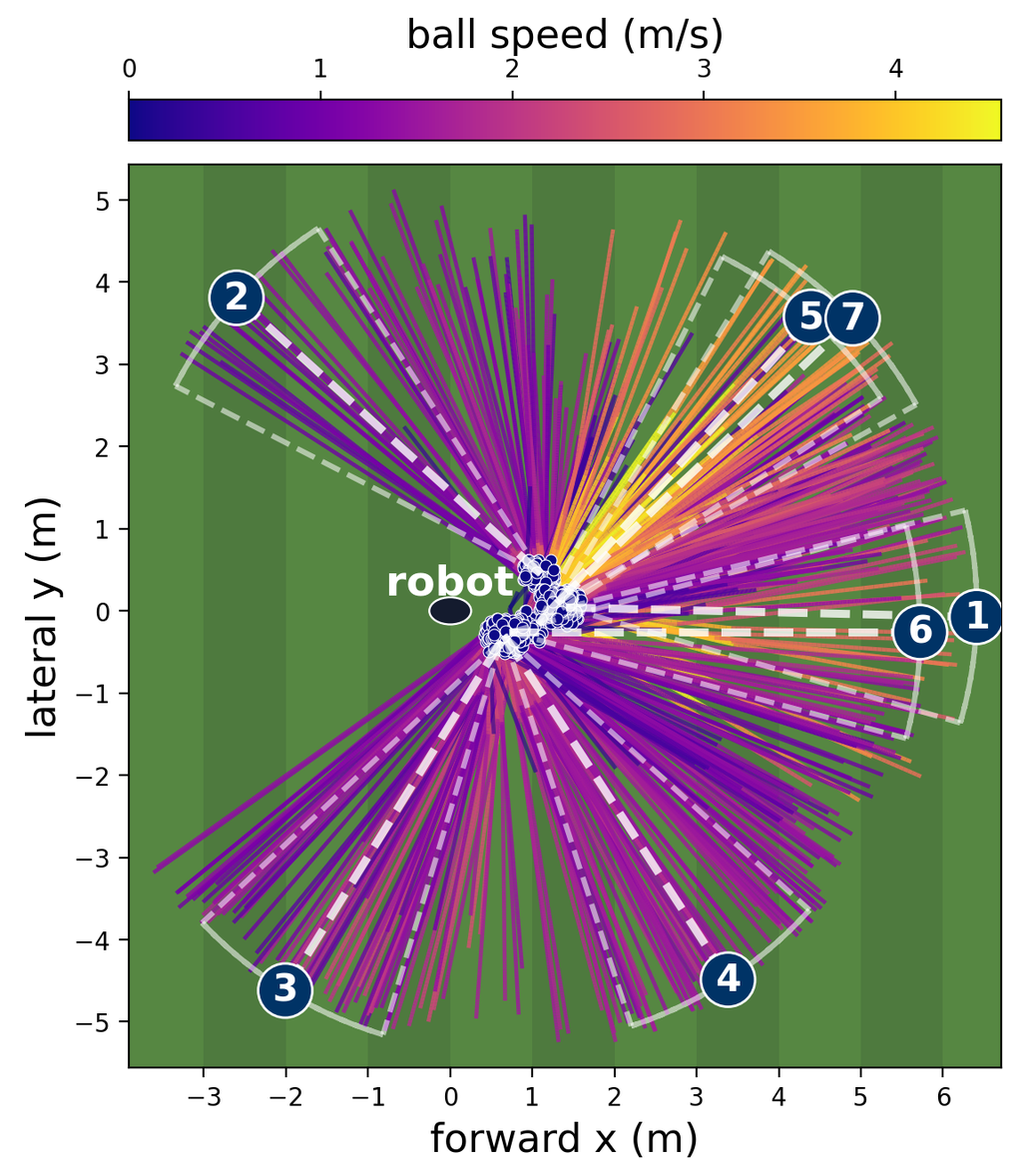}
    % \caption{Strike direction and pace, by skill.}
    \caption{Strike direction}
    \label{fig:coverage}
  \end{subfigure}
  \hfill
  \begin{subfigure}{0.49\columnwidth}
    \centering
    \includegraphics[width=\linewidth]{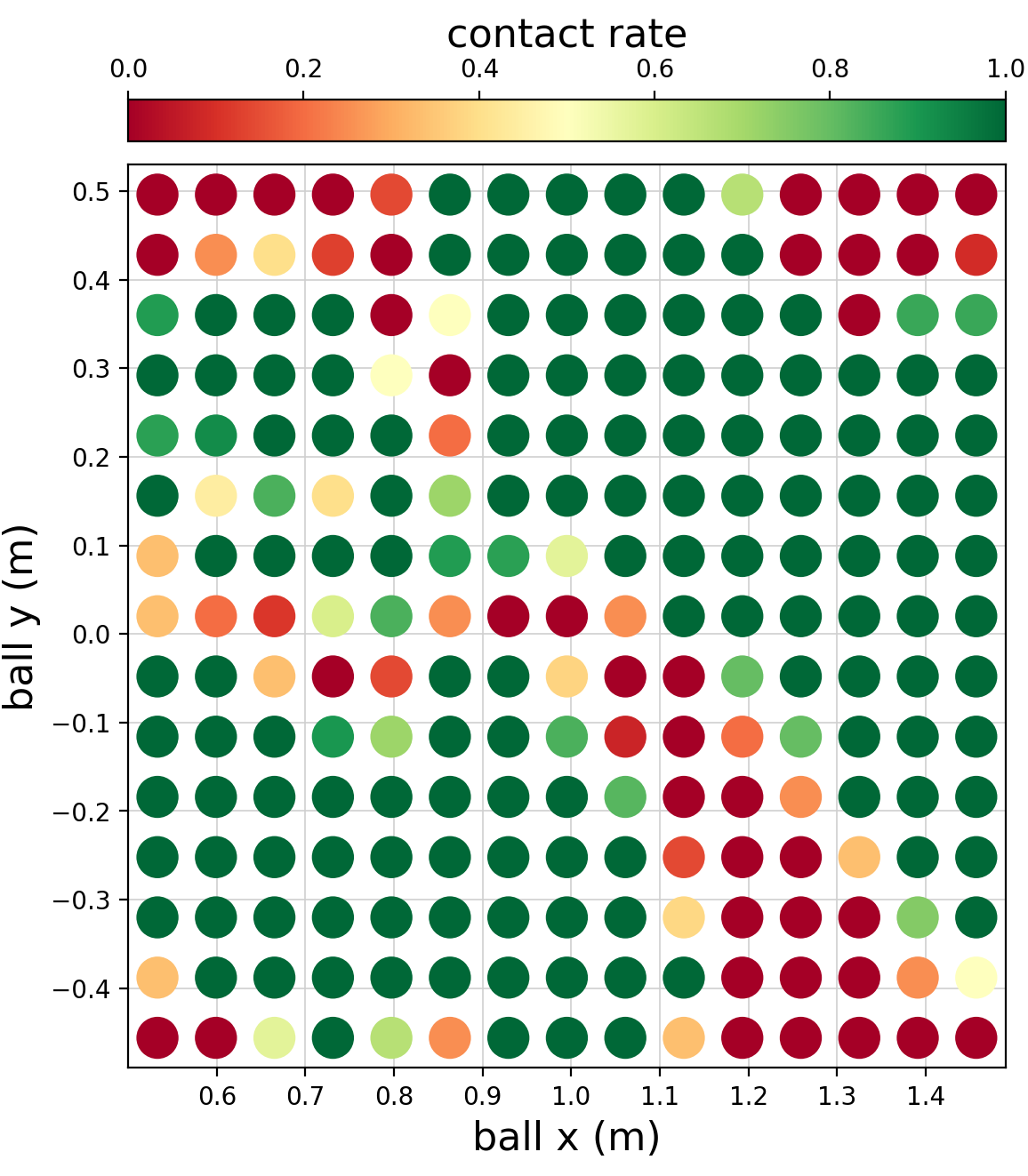}
    % \caption{Contact under displaced ball placement.}
    \caption{Contact under displacement}
    \label{fig:hitmap}
  \end{subfigure}

  \caption{\textbf{What the library covers, and how far the ball may sit
  from where a skill expects it.}}
  \label{fig:polar}
\end{figure}

Table~\ref{tab:perskill} reports the final unified policy's per-skill
performance, medians over $16$ checkpoints; Fig.~\ref{fig:polar} shows the
library geometrically. Panel (a) plots every ball contact from $100$ trials
per skill ($564$ of $700$ attempts; ball jittered $\pm20$\,cm, twice the
training range). The seven skills resolve into five islands, with skills 1/6
and 5/7 overlapping. Post-contact ball speed varies twofold
($2.04$--$4.13$\,m/s, Table~\ref{tab:perskill}) with no forward-versus-lateral
pattern. Trajectories spill outside their $\pm15^\circ$ aim bands, which is
the aim bias Sec.~\ref{sec:limitations} quantifies rather than commanded
spread. Panel (b) sweeps $1500$ placements over a region spanning all seven
spawns, assigning each to its nearest skill ($70.4\%$ contact overall). Of
these, $78.5\%$ lie outside every skill's $\pm10$\,cm training box. Contact
survives displacement far better than aim does: only $11.9\%$ of placements
pass within $1.0$\,m of the target.

Topple rates stay low across all seven bearings (mean $0.6\%$, worst
$2.6\%$), but aim does not: success spans $0.8\%$ to $36.6\%$, and
\textbf{interference from merging skills is the dominant cause.}
Table~\ref{tab:negtransfer} compares each skill's success against the
teacher it was distilled from, across four cumulative library sizes. Mean
success declines monotonically from $26.0\%$ at $N{=}2$ to $15.4\%$ at
$N{=}7$, a deficit already present at the smallest library and widening with
every skill added. The merge is therefore the bottleneck: every specialist was trained
from the same Stage-1 base. skill\_2 is the clearest case: distilled from one of the
weakest teachers ($19.6\%$) into the most extreme rotation ($+137.8^\circ$),
it contacts the ball $92.0\%$ of the time yet collapses to a $0.8\%$ median.
skill\_1 and skill\_6 contact on every attempt yet land only $10.2\%$ and
$7.6\%$. All three are aiming failures rather than striking ones: a bearing
correction alone lifts them to $20$--$29\%$ (Table~\ref{tab:calib}).

\begin{table}[htbp]
\centering
\caption{Per-skill performance of the unified policy.}
\label{tab:perskill}
\scriptsize
\setlength{\tabcolsep}{3pt}
\begin{tabular}{@{}lrrrrrrrr@{}}
\toprule
Skill & $1$ & $2$ & $3$ & $4$ & $5$ & $6$ & $7$ & mean \\
\midrule
Suc.@$0.5$ (\%) & $10.2$ & $0.8$ & $20.6$ & $25.8$ & $36.6$ & $7.6$ & $6.4$ & $15.4$ \\
Suc.@$1.0$ (\%) & $22.0$ & $3.2$ & $40.4$ & $51.0$ & $67.4$ & $23.8$ & $17.8$ & $32.2$ \\
Shot err.\ (m)  & $1.90$ & $2.92$ & $0.98$ & $1.10$ & $0.81$ & $1.89$ & $2.15$ & $1.68$ \\
$v^{\max}_{\mathrm{ball}}$ (m/s) & $3.63$ & $2.22$ & $2.84$ & $2.82$ & $2.04$ & $2.36$ & $4.13$ & $2.86$ \\
Hit (\%)        & $100.0$ & $92.0$ & $85.2$ & $100.0$ & $100.0$ & $100.0$ & $100.0$ & $96.7$ \\
Alive (\%)      & $99.6$ & $100.0$ & $98.6$ & $100.0$ & $100.0$ & $100.0$ & $97.4$ & $99.4$ \\
\bottomrule
\end{tabular}
\footnotesize
\setlength{\tabcolsep}{4pt}

\end{table}

\begin{table}[htbp]
\centering
\caption{Negative transfer: specialist vs.\ unified.}
\label{tab:negtransfer}
\scriptsize
\setlength{\tabcolsep}{2.5pt}
\begin{tabular}{@{}lccccc@{}}
\toprule
Skill & Teacher & $N{=}2$ & $N{=}4$ & $N{=}6$ & $N{=}7$ \\
\midrule
skill\_1 & $44.3$ & $37.8\,(-6.5)$ & $22.5\,(-21.8)$ & $29.0\,(-15.3)$ & $10.2\,(-34.1)$ \\
skill\_2 & $19.6$ & $14.2\,(-5.4)$ & $8.0\,(-11.6)$  & $5.3\,(-14.3)$  & $0.8\,(-18.8)$ \\
skill\_3 & $31.8$ & --             & $31.8\,(+0.0)$  & $19.2\,(-12.6)$ & $20.6\,(-11.2)$ \\
skill\_4 & $30.0$ & --             & $20.7\,(-9.3)$  & $8.8\,(-21.2)$  & $25.8\,(-4.2)$ \\
skill\_5 & $46.4$ & --             & --              & $30.2\,(-16.2)$ & $36.6\,(-9.8)$ \\
skill\_6 & $41.8$ & --             & --              & $15.3\,(-26.5)$ & $7.6\,(-34.2)$ \\
skill\_7 & $18.4$ & --             & --              & --              & $6.4\,(-12.0)$ \\
\midrule
mean     & $33.2$ & $26.0\,(-5.9)$ & $20.8\,(-10.7)$ & $18.0\,(-17.7)$ & $15.4\,(-17.8)$ \\
\bottomrule
\end{tabular}

\end{table}

\subsection{Handoff}
\label{sec:handoff-results}

Table~\ref{tab:handoff} evaluates the final unified policy (medians over
$16$ checkpoints) on both directions of the mode boundary.
Kick-to-locomotion, scored on survival alone, survives both the policy's
own auto-return and an abort at an arbitrary tick, including inside the
strike window. Locomotion-to-kick sweeps entry speed from zero to the full
locomotion range and is scored on survival, contact and aim, since contact
alone saturates near $100\%$ on most skills. \textbf{Entering the kick
while moving costs both stability and accuracy, and the cost grows smoothly
with entry speed.} Scaling the envelope moves contact $95.3{\to}84.1$,
success at $0.5$\,m $17.0{\to}11.9$, at $1.0$\,m $32.7{\to}25.7$, and topple
$0.7{\to}4.2$, monotonically on all four. Survival is not the fragile part: $95.8\%$ of full-speed entries survive,
against $99.3\%$ from rest. Accuracy is: success at $0.5$\,m
falls by nearly a third. The approach itself is unmeasured, since the ball is placed at its spawn rather than walked to.

Table~\ref{tab:ablations} isolates the two mechanisms behind this robustness
on skill\_7 specialist checkpoints, not the deployed library.
(a) ablates post-strike stabilisation, scored on settled-stance error rather
than survival, since all three arms already survive at or above $99\%$. Our
arm holds the lowest error and is the only one that settles at all.
Contact safety without the posture term is worse than neither, so the two are
not separable. (b) ablates the kick-state pose injection that trains
locomotion recovery from live, kick-disturbed states rather than idealised
reference-clip poses. The gap between an arm's ref.-clip and live-donor
columns widens monotonically the further training sits from the live
distribution: $6.6$ points for the transplant arm, $22.6$ for
reference-clip-only, $39.7$ for none. The transplant arm's live-donor $81.8\%$ is not
comparable to Table~\ref{tab:handoff}'s pooled $98.8\%$. Measured the same
strike-window-only way on the deployed checkpoints it is $98.1\%$, so the gap
is checkpoint maturity, not disagreement. Table~\ref{tab:hardware}(b) confirms the main result on
hardware: kick-to-locomotion survival under a mid-swing abort, triggered
manually and scored phase-stratified rather than by a scripted random
tick.

\begin{table}[htbp]
\centering
\caption{Handoff robustness, both directions.}
\label{tab:handoff}
\footnotesize
% 2.5pt, not the 4pt this table used at four columns: the two success columns added 2026-09-12
% push the tabular 13.9pt past \columnwidth at 4pt. Tightening the gutters buys 18pt and keeps
% every label unabbreviated, which is preferable to shortening the trigger-condition wording.
\setlength{\tabcolsep}{2.5pt}
\begin{tabular}{@{}llcccc@{}}
\toprule
Direction & Trigger condition & Alive & Hit & Suc.@$0.5$ & Suc.@$1.0$ \\
\midrule
Kick$\to$Loco & end of authored clip       & $99.1$  & n/a & n/a & n/a \\
Kick$\to$Loco & random tick, incl.\ strike & $98.8$ & n/a & n/a & n/a \\
\addlinespace[1pt]
Loco$\to$Kick & from $v\!\approx\!0$ state  & $99.3$ & $95.3$ & $17.0$ & $32.7$ \\
Loco$\to$Kick & in motion, $25\%$ range & $97.2$ & $90.2$ & $14.7$ & $29.8$ \\
Loco$\to$Kick & in motion, $50\%$ range & $97.0$ & $87.8$ & $13.7$ & $27.5$ \\
Loco$\to$Kick & in motion, full range & $95.8$ & $84.1$ & $11.9$ & $25.7$ \\
\bottomrule
\end{tabular}
\end{table}

\begin{table}[t]
\centering
\caption{Hardware results.}
\label{tab:hardware}
\scriptsize
\setlength{\tabcolsep}{3pt}

\textbf{(a) Shooting trials.}\\[2pt]
\begin{tabular}{@{}lccccc@{}}
\toprule
Condition & Trials & Suc.$^\dagger$ (\%) & Err.\ (m) & Hit (\%) & Alive (\%) \\
\midrule
Static ball, forward  & $32$ & $46.9$ & $0.25$ & $93.8$ & $100.0$ \\
Static ball, lateral  & $30$ & $53.3$ & $0.19$ & $90.0$ & $100.0$ \\
Static ball, rearward & $30$ & $20.0$ & $0.87$ & $93.3$ & $96.7$  \\
Commanded approach    & $30$ & $40.0$ & $0.30$ & $93.3$ & $100.0$ \\
\bottomrule
\end{tabular}

{\scriptsize $^{\dagger}$Radius $0.14$\,m, i.e.\ $\pm5.7^\circ$ at $1.42$\,m,
not the simulated $0.5$\,m.}

\vspace{2pt}
\textbf{(b) Handoff robustness, Kick$\to$Loco, manual abort ($20$ trials
each).}\\[2pt]
\begin{tabular}{@{}lccc@{}}
\toprule
Abort phase & Pre-strike & Mid (contact) & Post/recovery \\
\midrule
Alive (\%) & $95.0$ & $95.0$ & $100.0$ \\
\bottomrule
\end{tabular}

\end{table}

\begin{table}[htbp]
\centering
\caption{Ablations on skill\_7, medians over eight matched Stage-2
checkpoints per arm.}
\label{tab:ablations}
\footnotesize
\setlength{\tabcolsep}{4pt}

\textbf{(a) Post-strike stabilisation mechanism.}\\[2pt]
\scriptsize
\setlength{\tabcolsep}{3pt}
\begin{tabular}{@{}p{3.15cm}ccc@{}}
\toprule
Arm & \makecell{Stance\\asym.$\downarrow$\\[-1pt]{\tiny rad$^2$}}
    & \makecell{Yaw\\drift$\downarrow$\\[-1pt]{\tiny (rad/s)$^2$}}
    & \makecell{Orient.\\tilt$\downarrow$\\[-1pt]{\tiny $\sin\theta$}} \\
\midrule
Posture + contact safety (ours) & $0.013$ & $0.012$ & $0.003$ \\
Contact safety only             & $0.276$ & $0.074$ & $0.163$ \\
No stabilisation                & $0.117$ & $0.044$ & $0.073$ \\
\bottomrule
\end{tabular}
\setlength{\tabcolsep}{4pt}

\vspace{2pt}
\textbf{(b) Kick-state pose injection for locomotion recovery.}\\[2pt]
\scriptsize
\setlength{\tabcolsep}{3pt}
\begin{tabular}{@{}p{3.1cm}cc@{}}
\toprule
Training arm & \makecell{Alive,\\ref.-clip (\%)} & \makecell{Alive,\\live donor (\%)} \\
\midrule
Transplant + ref.-clip (ours) & $88.4$ & $81.8$ \\
Reference-clip pose only      & $82.3$ & $59.7$ \\
No pose injection             & $71.8$ & $32.1$ \\
\bottomrule
\end{tabular}

\vspace{2pt}
\textbf{(c) Locomotion-mode tracking gate.}\\[2pt]
\begin{tabular}{@{}lcccc@{}}
\toprule
 & \multicolumn{2}{c}{$v_x$ (m/s)} & $v_y$ & $\omega_z$ \\
\cmidrule(lr){2-3}
Tracking MAE$\downarrow$ & fwd. & back. & (m/s) & (rad/s) \\
\midrule
Hard zero (ours) & $0.34$ & $0.46$ & $0.35$ & $0.14$ \\
Floor $0.1$      & $0.38$ & $0.58$ & $0.37$ & $0.15$ \\
Floor $0.5$      & $0.67$ & $0.79$ & $0.40$ & $0.17$ \\
\bottomrule
\end{tabular}
\footnotesize
\setlength{\tabcolsep}{4pt}

\end{table}

\subsection{Locomotion generality}

Table~\ref{tab:loco} is the axis prior systems do not report: paired against
the same substrate trained without any skill layer, the claim is an absence
of effect: attaching seven skills should not degrade locomotion. We retrained the substrate under two
further seeds to give that claim a noise floor: every $\Delta$ sits inside one
standard deviation of the baseline's own three-seed spread, and all but
backward tracking inside half of it. Leaving the library side at one seed only understates that floor, but it
cannot exclude a small degradation reproducing across seeds: all four tracking
deltas are non-negative. Terrain traversal
uses the same sustained-fall criterion training itself uses ($10$
consecutive ticks below threshold, not an instantaneous dip).

\begin{table}[htbp]
\centering
\caption{Locomotion generality with the skill library attached. Both policies
build on the seed-$42$ substrate, and $\Delta$ is that paired difference,
computed before rounding. The s.d.\ row is the spread of three Stage-1 seeds
($42/7/13$), a noise floor rather than an interval around the row above:
their terrain rates are $98.7/97.3/62.7$.}
\label{tab:loco}
\scriptsize
\setlength{\tabcolsep}{3pt}
\begin{tabular}{@{}lcccccc@{}}
\toprule
 & \multicolumn{2}{c}{$v_x$ MAE} & $v_y$ & $\omega_z$ & Terr. & Push \\
\cmidrule(lr){2-3}
 & fwd. & back. & \multicolumn{2}{c}{\tiny (m/s), (rad/s)} & (\%) & (\%) \\
\midrule
Loco-only, seed $42$ & $0.39$ & $0.51$ & $0.35$ & $0.21$ & $98.7$ & $100.0$ \\
\;\;s.d.\ over $3$ seeds & $0.07$ & $0.05$ & $0.07$ & $0.17$ & $20.4$ & $0.0$ \\
With library & $0.42$ & $0.55$ & $0.35$ & $0.25$ & $100.0$ & $100.0$ \\
$\Delta$ & $+0.02$ & $+0.04$ & $+0.00$ & $+0.04$ & $+1.3$ & $+0.0$ \\
\bottomrule
\end{tabular}

\end{table}

\section{LIMITATIONS AND CONCLUSION}
\label{sec:limitations}

\subsection{Limitations}

\textbf{Distillation loses most of the teachers' accuracy, and the loss grows
with library size.} This is the system's principal open problem
(Table~\ref{tab:negtransfer}). Mean success at $0.5$\,m falls from $33.2\%$
across teachers to $15.4\%$ merged. Whether this is a capacity limit or a fixable
distillation artifact is open: skills move non-monotonically across library
sizes, and the distillation has no seed replicates.
Our multi-skill claim is thus about coverage and stability, not specialist
accuracy. The gate ablation is likewise one skill at Stage 2, one seed. It shows that
discounting tracking costs command tracking, not that the hard zero is
required at library scale.

\textbf{The strikes are both mis-aimed and imprecise.} Across $112$
checkpoint-skill regressions of the ball's post-contact bearing on the
commanded offset $\theta$, per-skill median slopes span $0.02$ to $0.88$
rather than $1$, and one fit in five is negative, steering the wrong way. Each fit also carries an
angular offset, up to $40^\circ$ by skill and unstable across checkpoints.
Per-trial scatter about that offset is $9$--$24^\circ$, wider than the
$\pm5.7^\circ$ a $0.5$\,m hit allows. Both terms therefore bound the
shot-error and success columns of Tables~\ref{tab:context}
and~\ref{tab:perskill}.
Correcting the offset recovers much of this
(Table~\ref{tab:calib}): refitting each skill's nominal bearing to its
measured departure bearing, cross-validated within each checkpoint, improves
success at both radii by $1.7$--$1.8\times$ over the same trials scored
without it, on all $16$ checkpoints (sign test $p<10^{-4}$) and every skill. The command is a relative offset, so this is what a
recalibrated deployment would measure, not a projection. The offsets replicate across halves of a checkpoint but not across
checkpoints, so the fit is per deployed checkpoint, not once per skill.

\begin{table}[htbp]
\centering
\caption{Per-skill angular calibration, medians over Table~\ref{tab:perskill}'s
$16$ checkpoints, on a separate $250$-trial scan. $\delta$ is fitted on half
of each checkpoint's trials and scored on the other, then swapped.}
\label{tab:calib}
\scriptsize
\setlength{\tabcolsep}{3pt}
\begin{tabular}{@{}lrrrrrrrr@{}}
\toprule
Skill & $1$ & $2$ & $3$ & $4$ & $5$ & $6$ & $7$ & All \\
\midrule
$\delta$ (deg)        & $+21$ & $-40$ & $-8$ & $-11$ & $-5$ & $-17$ & $0$ & -- \\
\midrule
Suc.@$0.5$ (\%)       & $10.4$ & $1.2$ & $21.8$ & $26.8$ & $36.2$ & $8.6$ & $9.6$ & $16.4$ \\
\;\;+ calibration     & $\mathbf{28.8}$ & $\mathbf{20.2}$ & $\mathbf{29.0}$ & $\mathbf{41.6}$ & $\mathbf{40.4}$ & $\mathbf{27.4}$ & $\mathbf{16.8}$ & $\mathbf{29.2}$ \\
\midrule
Suc.@$1.0$ (\%)       & $22.4$ & $4.8$ & $38.8$ & $52.2$ & $65.2$ & $24.0$ & $20.2$ & $32.5$ \\
\;\;+ calibration     & $\mathbf{56.0}$ & $\mathbf{42.4}$ & $\mathbf{49.0}$ & $\mathbf{71.6}$ & $\mathbf{74.8}$ & $\mathbf{51.2}$ & $\mathbf{33.6}$ & $\mathbf{54.1}$ \\
\bottomrule
\end{tabular}
\end{table}

\subsection{Conclusion}

We presented a humanoid soccer policy built the other way around from current
practice. A commandable locomotion policy is trained first and alone, with $N$
motion-guided kicking skills layered on as task-gated additions held at
exactly zero outside kick mode. The hard-zero gate fixes the reachable gait
space to the command curriculum, not to any reference clip. And because every
skill begins and ends in the same locomotion state, a library of $N$ skills
needs $O(N)$ transitions, not $O(N^2)$.

On the metrics prior systems report, the locomotion-grounded specialist
reaches $44.3\%$ success at $0.5$\,m, over a narrower aim cone and at lower
pace than prior work. The distilled library averages $15.4\%$ across the seven
bearings, at a locomotion cost of at most $0.04$ in command-tracking MAE, and
transfers to hardware.

% Flush every pending float here: no table may appear after the references.
\FloatBarrier

% \addtolength{\textheight}{-2cm}   % Balances the column lengths on the last page.

%%%%%%%%%%%%%%%%%%%%%%%%%%%%%%%%%%%%%%%%%%%%%%%%%%%%%%%%%%%%%%%%%%%%%%%%%%%%%%%%

\bibliographystyle{IEEEtran}
\bibliography{references}

\begin{thebibliography}{10}
\providecommand{\url}[1]{#1}
\csname url@samestyle\endcsname
\providecommand{\newblock}{\relax}
\providecommand{\bibinfo}[2]{#2}
\providecommand{\BIBentrySTDinterwordspacing}{\spaceskip=0pt\relax}
\providecommand{\BIBentryALTinterwordstretchfactor}{4}
\providecommand{\BIBentryALTinterwordspacing}{\spaceskip=\fontdimen2\font plus
\BIBentryALTinterwordstretchfactor\fontdimen3\font minus
  \fontdimen4\font\relax}
\providecommand{\BIBforeignlanguage}[2]{{%
\expandafter\ifx\csname l@#1\endcsname\relax
\typeout{** WARNING: IEEEtran.bst: No hyphenation pattern has been}%
\typeout{** loaded for the language `#1'. Using the pattern for}%
\typeout{** the default language instead.}%
\else
\language=\csname l@#1\endcsname
\fi
#2}}
\providecommand{\BIBdecl}{\relax}
\BIBdecl

\bibitem{haarnoja2024soccer}
T.~Haarnoja \emph{et~al.}, ``Learning agile soccer skills for a bipedal robot
  with deep reinforcement learning,'' \emph{Sci. Robot.}, vol.~9, no.~89, p.
  eadi8022, 2024.

\bibitem{robonaldo}
Y.~Zhong \emph{et~al.}, ``{RoboNaldo}: Accurate, stable and powerful humanoid
  soccer shooting via motion-guided curriculum reinforcement learning,''
  \emph{arXiv:2606.11092}, 2026.

\bibitem{paid}
J.~Kong \emph{et~al.}, ``Learning soccer skills for humanoid robots: A
  progressive perception-action framework,'' \emph{arXiv:2602.05310}, 2026.

\bibitem{striker}
Z.~Xu, M.~Seo, D.~Lee \emph{et~al.}, ``Learning agile striker skills for
  humanoid soccer robots from noisy sensory input,'' in \emph{Proc. IEEE ICRA},
  2026.

\bibitem{humanx}
Y.~Wang \emph{et~al.}, ``{HumanX}: Toward agile and generalizable humanoid
  interaction skills from human videos,'' \emph{arXiv:2602.02473}, 2026.

\bibitem{reactive}
------, ``Learning vision-driven reactive soccer skills for humanoid robots,''
  \emph{arXiv:2511.03996}, 2025.

\bibitem{goalkeeper}
J.~Ren \emph{et~al.}, ``Humanoid goalkeeper: Learning from position conditioned
  task-motion constraints,'' \emph{arXiv:2510.18002}, 2025.

\bibitem{gvhmr}
Z.~Shen, H.~Pi, Y.~Xia \emph{et~al.}, ``World-grounded human motion recovery
  via gravity-view coordinates,'' in \emph{Proc. SIGGRAPH Asia}, 2024.

\bibitem{gmr}
J.~P. Ara{\'u}jo \emph{et~al.}, ``Retargeting matters: General motion
  retargeting for humanoid motion tracking,'' \emph{arXiv:2510.02252}, 2025.

\bibitem{deepmimic}
X.~B. Peng \emph{et~al.}, ``{DeepMimic}: Example-guided deep reinforcement
  learning of physics-based character skills,'' \emph{ACM Trans. Graph.},
  vol.~37, no.~4, pp. 1--14, 2018.

\bibitem{amp}
------, ``{AMP}: Adversarial motion priors for stylized physics-based character
  control,'' \emph{ACM Trans. Graph.}, vol.~40, no.~4, pp. 1--20, 2021.

\bibitem{beyondmimic}
Q.~Liao \emph{et~al.}, ``{BeyondMimic}: From motion tracking to versatile
  humanoid control via guided diffusion,'' \emph{arXiv:2508.08241}, 2025.

\bibitem{isaaclab}
M.~Mittal \emph{et~al.}, ``{Orbit}: A unified simulation framework for
  interactive robot learning environments,'' \emph{IEEE Robot. Autom. Lett.},
  vol.~8, pp. 3740--3747, 2023.

\bibitem{sac}
T.~Haarnoja \emph{et~al.}, ``Soft actor-critic: Off-policy maximum entropy deep
  reinforcement learning with a stochastic actor,'' in \emph{Proc. ICML}, 2018,
  pp. 1861--1870.

\bibitem{fastsac}
Y.~Seo \emph{et~al.}, ``Learning sim-to-real humanoid locomotion in 15
  minutes,'' \emph{arXiv:2512.01996}, 2025.

\bibitem{dagger}
S.~Ross, G.~J. Gordon, and J.~A. Bagnell, ``A reduction of imitation learning
  and structured prediction to no-regret online learning,'' in \emph{Proc.
  AISTATS}, 2011, pp. 627--635.

\bibitem{policydistillation}
A.~A. Rusu, S.~G. Colmenarejo, C.~Gulcehre \emph{et~al.}, ``Policy
  distillation,'' in \emph{Proc. ICLR}, 2016.

\bibitem{mujoco}
E.~Todorov, T.~Erez, and Y.~Tassa, ``{MuJoCo}: A physics engine for model-based
  control,'' in \emph{Proc. IEEE/RSJ IROS}, 2012, pp. 5026--5033.

\bibitem{stoft}
W.~Li \emph{et~al.}, ``Like playing a video game: Spatial-temporal optimization
  of foot trajectories for controlled football kicking in bipedal robots,'' in
  \emph{Proc. IEEE/RSJ IROS}, 2025, pp. 3565--3572.

\bibitem{skillx}
Z.~Ye \emph{et~al.}, ``{SkillX}: Unified multi-skill policy learning for
  humanoid soccer,'' \emph{arXiv:2609.06718}, 2026.

\bibitem{falcon}
Y.~Zhang \emph{et~al.}, ``{FALCON}: Learning force-adaptive humanoid
  loco-manipulation,'' in \emph{Proc. L4DC}, 2026.

\bibitem{options}
R.~S. Sutton, D.~Precup, and S.~Singh, ``Between {MDP}s and semi-{MDP}s: A
  framework for temporal abstraction in reinforcement learning,'' \emph{Artif.
  Intell.}, vol. 112, no. 1-2, pp. 181--211, 1999.

\bibitem{ng1999shaping}
A.~Y. Ng, D.~Harada, and S.~Russell, ``Policy invariance under reward
  transformations: Theory and application to reward shaping,'' in \emph{Proc.
  ICML}, 1999, pp. 278--287.

\bibitem{ppo}
J.~Schulman \emph{et~al.}, ``Proximal policy optimization algorithms,''
  \emph{arXiv:1707.06347}, 2017.

\end{thebibliography}

\end{document}